\documentclass[11pt]{article}

\usepackage[final]{acl}  

\usepackage{times}
\usepackage{latexsym}
\usepackage[T1]{fontenc}
\usepackage[utf8]{inputenc}
\usepackage{microtype}
\usepackage{inconsolata}
\usepackage{graphicx}
\usepackage{booktabs}
\usepackage{amsmath}
\usepackage{amssymb}
\usepackage{multirow}
\usepackage[table]{xcolor}
\usepackage{subcaption}
\usepackage{tcolorbox}
\usepackage{makecell}
\usepackage{algorithm}
\usepackage{algorithmic}
\usepackage{pifont}
\usepackage{enumitem}
\usepackage{xspace}
\newcolumntype{M}[1]{>{\centering\arraybackslash}m{#1}}

\newcommand{\ourmodel}{{\texttt{\textbf{HiPS}}}\xspace}
\definecolor{cvprblue}{RGB}{31,119,180}
\definecolor{rowno}{RGB}{245,245,245}       
\definecolor{rowwf}{RGB}{232,240,254}       
\definecolor{rowrl}{RGB}{232,248,232}       
\definecolor{rowours}{RGB}{255,249,230}     

\newcommand{\Su}{S_u}
\newcommand{\Dp}{\Delta_p}
\newcommand{\Sp}{S_p}
\newcommand{\Rans}{R_{\text{ans}}}
\newcommand{\Rfollow}{R_{\text{follow}}}
\newcommand{\PG}{\mathrm{PG}}
\newcommand{\Diverg}{D}

\title{Learning What to Share and What to Personalize: \\ Hierarchical Strategy Co-Evolution for Agent Memory}

\author{
{\setlength{\tabcolsep}{4pt}
\begin{tabular}{@{}cccccc@{}}
Yupeng Han &
Shuochen Liu &
Kai Zhang\thanks{Corresponding author.} &
Ze Liu &
Zhihong Pan &
Xianquan Wang
\end{tabular}}
\\[2mm]
State Key Laboratory of Cognitive Intelligence, \\
University of Science and Technology of China
\\
\texttt{yupenghan@mail.ustc.edu.cn} \quad
\texttt{kkzhang08@ustc.edu.cn}
}

\begin{document}
\maketitle

\begin{abstract}
Memory-augmented agents maintain compact user profiles throughout extended conversations, enabling personalized and consistent responses without the need to process the entire dialogue history. The quality of these user profiles relies on the underlying memory management strategy: at each step, the agent must determine what to retain, compress, or discard. However, existing methods typically employ a static, one-size-fits-all strategy established before training. In practice, the optimal memory decision is inherently user-specific and dynamically evolves alongside policy optimization. To address this, we propose \textbf{HiPS} (\textbf{Hi}erarchical \textbf{P}ersonalized \textbf{S}trategy), a framework that decouples memory management into a globally shared foundation and a user-specific adaptive tier. Specifically, HiPS employs \textbf{Universal Strategy} to extract shared principles from cross-persona trajectories, alongside \textbf{Persona Delta Distillation} to generate tailored rules for users whose behaviors diverge from general patterns. \textbf{Cross-Level Rule Flow} dynamically calibrates their boundary by promoting broadly validated personal rules and demoting contradicted global ones. The architecture establishes a co-evolution loop where a mechanism guarantees that all strategy refinements are anchored to task outcomes. Extensive experiments demonstrate consistent improvements over memory-augmented baselines\footnote{https://github.com/Hyp26cs/HiPS}.

\end{abstract}

\begin{figure}
\centering
\includegraphics[width=\linewidth]{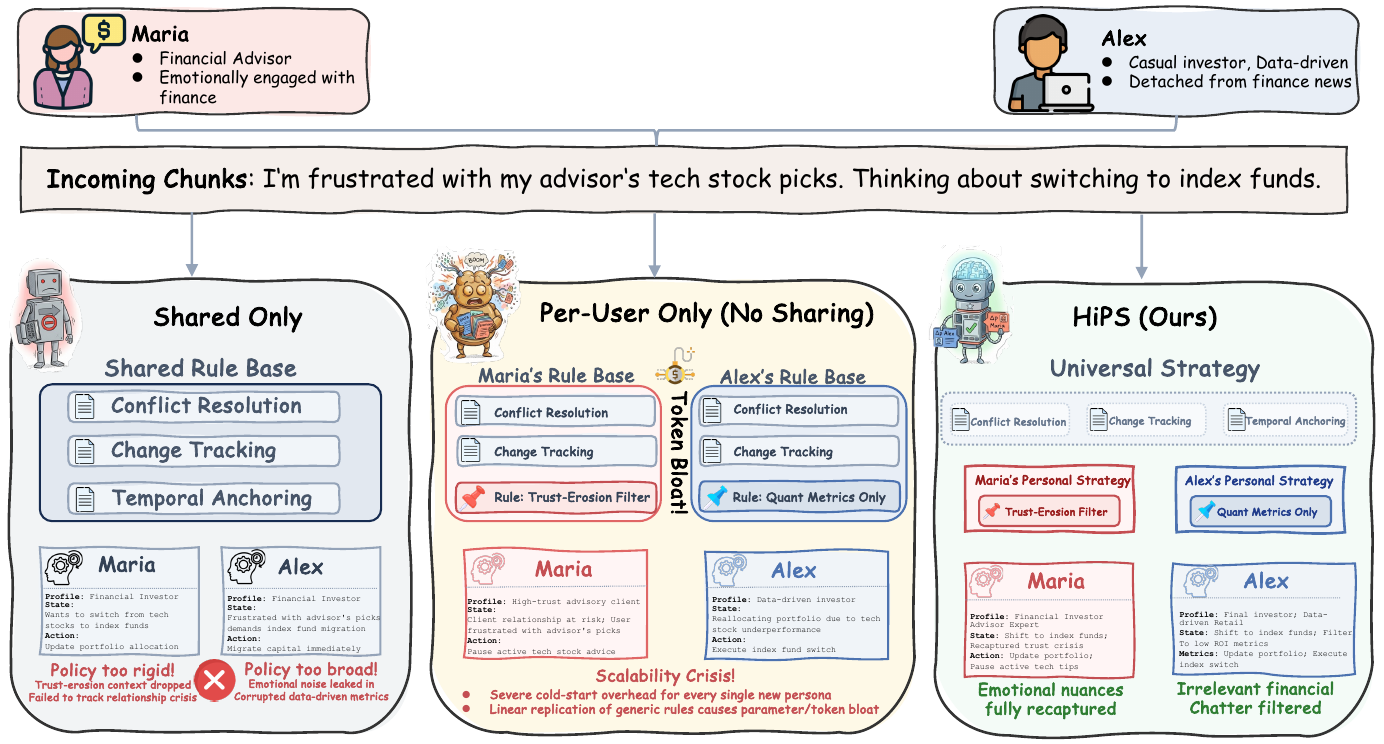}
\caption{Same conversation, different users, different needs. A shared strategy (left) applies identical rules to all users, losing Maria's emotional context while over-preserving Alex's trivial details. A hypothetical per-user independent approach (center) would redundantly rediscover universal rules and fail on cold-start users. HiPS (right) decomposes the strategy into a shared level and per-user adaptive level.}
\label{fig:motivation}
\vspace{-0.7cm}
\end{figure}
\section{Introduction}


Large language models (LLMs) are increasingly deployed as personalized conversational agents, which require maintaining coherent and tailored interactions for individual users~\cite{personamem,prefeval}. However, the inherent constraints of finite context windows prohibit LLMs from retaining unbounded dialogue histories~\cite{amem,gao-etal-2025-efficient-context}. Moreover, naive storage and retrieval of raw dialogue fail to capture the dynamic, evolving user preferences~\cite{rfmem}. This challenge motivates the integration of external memory systems, requiring the agent to dynamically evaluate at each turn whether to retain, compress, or discard incoming context.

Along this line, most existing memory systems employ static workflows, converting raw dialogue into external memory via predefined extraction and compression rules~\cite{mem0,amem,lightmem}. However, they lack the capacity to learn from interaction feedback or adapt to diverse user behaviors. Recent approaches formalize memory operations as learnable actions, optimizing update policies through reinforcement learning (RL)~\cite{memagent2025,agemem,mem1} or natural-language strategy distillation~\cite{memcoe2026}. While more adaptive, they impose a strictly user-agnostic management paradigm. When optimizing for population-averaged rewards, the learning signals for niche behaviors are overwhelmed by the majority~\cite{poddar2024personalizing}, resulting in a one-size-fits-all compromise. Moreover, methods utilizing explicit strategies~\cite{memcoe2026} keep these rules frozen during training. Blind to online feedback, the prescribed strategy misaligns with actual rollout trajectories as the policy evolves, making it imperative that strategies be both \emph{personalized} and \emph{adaptive}~\cite{survey}. Yet, individualizing every operation is redundant, as foundational rules are universally beneficial while niche behaviors require tailored handling. Since the optimal boundary is unknown a priori, this naturally raises the question: \textit{Can we dynamically discover and co-evolve this partition using on-policy evidence?}
To this end, we introduce \textbf{HiPS} (\textbf{Hi}erarchical \textbf{P}ersonalized \textbf{S}trategy), a framework that decomposes the management strategy into a shared universal level and a per-user adaptive level, dynamically learning the boundary between them from on-policy evidence. As illustrated in Fig.~\ref{fig:motivation}, HiPS contains three core mechanisms: (1) \textbf{Universal Strategy Distillation (USD)} evolves shared rules using cross-persona trajectories. By contrasting high- and low-reward episodes, USD generates structured revisions that validate or refine rules. (2) \textbf{Persona Delta Distillation (PDD)} formulates adaptive, per-user rules specifically for individuals whose behaviors diverge from the population norm. A divergence criterion selectively identifies users who need personalization, thereby preventing overfitting and noise for those adequately served by universal rules. Finally, (3) \textbf{Cross-Level Rule Flow} dynamically calibrates the boundary between these tiers by promoting broadly validated personal rules to the universal level and substituting contradicted global rules with targeted PDD revisions. Unlike previous approaches that freeze strategies before training \cite{memcoe2026}, HiPS interleaves strategy discovery and policy optimization within a unified, continuous loop. In this paradigm, the active strategy directs model rollouts to generate trajectories that subsequently serve as empirical feedback for further strategy refinement. To prevent self-reinforcing confirmation bias \cite{selfreward}, strategy evolution is anchored strictly to objective task outcomes, reserving rule adherence exclusively for policy optimization.
Our contributions are summarized as below:
\begin{itemize}[leftmargin=*, topsep=1pt, itemsep=0pt]
    \item We formulate the \textbf{strategy personalization problem}, demonstrating that the boundary between shared and user-specific memory rules must be discovered empirically rather than predefined.
    \item We propose \textbf{HiPS}, an RL framework that co-evolves a universal strategy and user-specific adaptive rules alongside a policy via evidence-based distillation and cross-level rule flow.
    \item Extensive experiments demonstrate consistent performance gains. We uncover a \textbf{component importance flip} where universal rules dominate in-domain tasks and adaptive mechanisms drive out-of-domain generalization.
\end{itemize}

\section{Related Work}

\paragraph{Memory architectures.}
LLM agent memory systems manage how historical information is stored, organized, and retrieved within a bounded context window. Memory bank approaches apply segmentation, summarization, and selective forgetting to maintain long-term quality~\cite{mem0,amem,lightmem}. Structured indices such as tree- and graph-based retrieval improve access efficiency~\cite{memgpt,memos2025}. Personalization-oriented systems extract and maintain user profiles for downstream conditioning~\cite{personamem}. Despite their diversity, these systems share two fundamental limitations: their management strategies are \textbf{fixed}, rendering them incapable of learning from interaction feedback, and \textbf{shared}, imposing identical rules across all users regardless of behavioral differences.

\begin{figure*}[t]
\centering
\includegraphics[width=\linewidth]{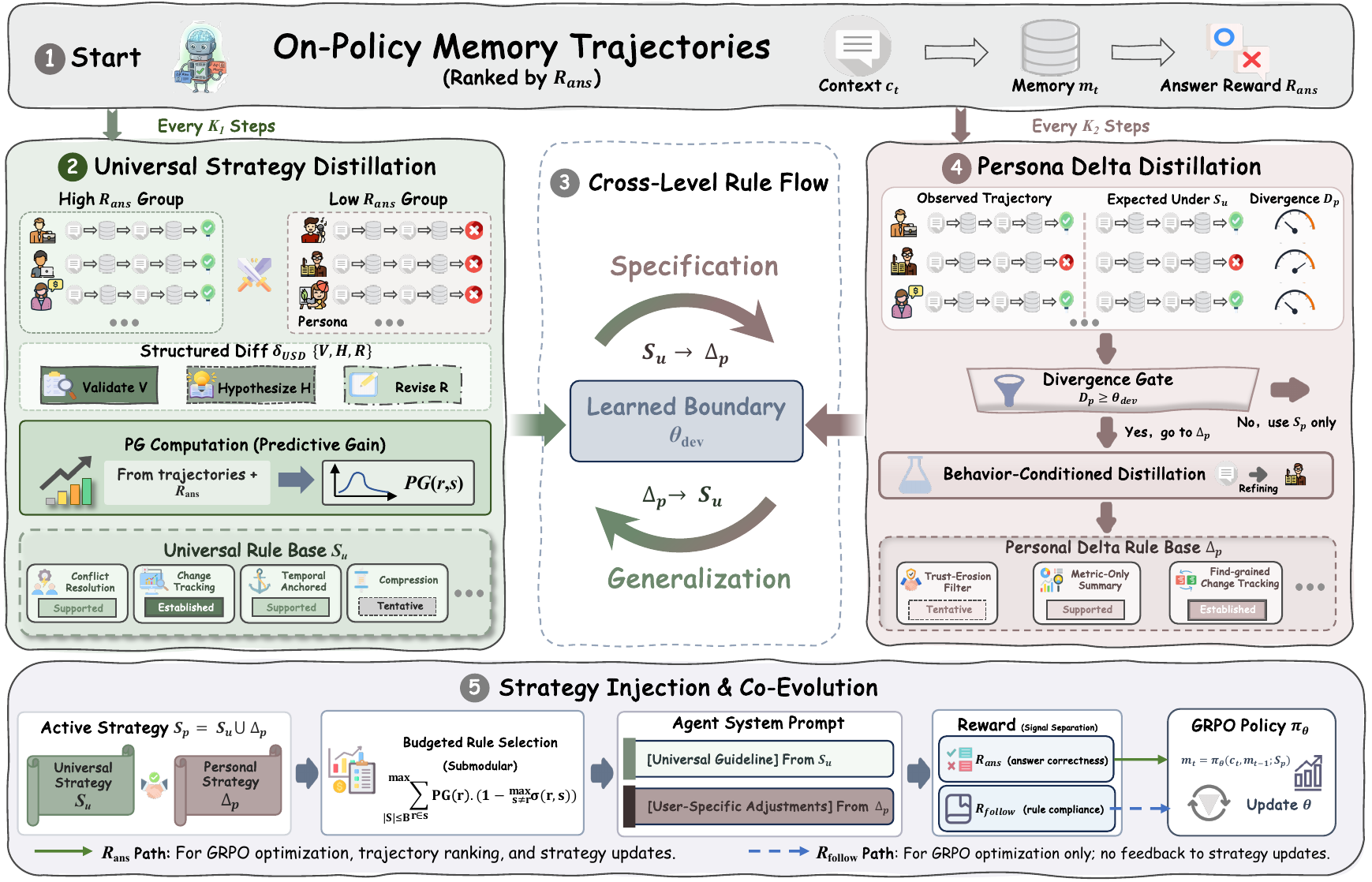}
\caption{HiPS framework overview. Universal Strategy Distillation (USD) evolves shared strategies $\Su$ from cross-persona trajectories. Persona Delta Distillation (PDD) evolves per-user adaptations $\Dp$ for users whose divergence exceeds $\theta_{\text{div}}$. Cross-Level Flow promotes widely validated deltas to $\Su$. The active strategy $S_p = \Su \cup \Dp$ is injected into the agent prompt via submodular selection.}
\label{fig:method}
\end{figure*}

\paragraph{RL-trained and strategy-based memory policies.}
To overcome the rigidity of fixed pipelines, recent studies formulate memory operations as learnable actions via reinforcement learning (RL)~\cite{memoryr1,memalpha,agemem,mem1,memagent2025}. While these methods effectively learn parametric and procedural \emph{policies}, they fail to develop declarative and interpretable \emph{strategies}: management behaviors are implicitly encoded within model parameters, rendering them neither inspectable nor editable. A parallel line of research formulates strategies as natural-language rules. For instance, MemCoE~\cite{memcoe2026} optimizes management guidelines using TextGrad before RL training, subsequently injecting them as system prompts for GRPO fine-tuning. EverMemOS~\cite{ever} introduces skill retirement through experience clustering and distillation, while MemSkill~\cite{memskill} reframes traditional static memory operations into learnable and evolvable ``memory skills'', dynamically optimizing both skill selection and refinement. Although these approaches yield interpretable strategies, the resulting rules are inherently \textbf{globally shared and frozen prior to training}: a single guideline serves all users and remains static once optimized. HiPS differs by employing \emph{user-indexed} strategies that dynamically \emph{co-evolve} with the policy driven by on-policy evidence, rather than relying on static, globally shared rules.

\section{Preliminary}

\label{sec:problem}

We consider a conversational setting in which a memory-augmented agent interacts with a user $p$ over multiple sessions. The dialogue history is segmented into contextual chunks $\{c_1, \ldots, c_K\}$, processed sequentially by the agent $\pi_\theta$. To facilitate long-term personalization, the agent maintains a compact memory state $m_t$ that is updated as new information arrives. The memory update mechanism is governed by a set of management rules, denoted as $\Sp$, which is injected into the agent's system prompt to regulate its actions at step $t$:
\begin{equation}
    m_t = \pi_\theta(c_t, m_{t-1}; \, \Sp).
\end{equation}

Prior work treats the management strategy as a single, globally shared entity that is fixed prior to training. This introduces two primary limitations: the strategy lacks user-specific adaptability, and it remains static as the policy evolves. To overcome these limitations and enable dynamic adaptation, we decompose $\Sp$ into shared and personalized components, allowing this partition to co-evolve with the policy during training. Formally,
\begin{equation}
    \Sp = \Su \cup \Dp,
\end{equation}
where $\Su$ comprises universally beneficial rules applicable to the entire population, and $\Dp$ contains adaptive rules tailored to the behavioral patterns of user $p$. After processing all $K$ dialogue chunks, the agent generates a final response to a query $q$ conditioned on the terminal memory state $m_K$. 

\section{Method}
In this section, we propose HiPS, a framework that decouples the memory management strategy into a shared universal tier and a user-specific adaptive tier. Specifically, Universal Strategy Distillation (USD) (Section~\ref{sec:usd}) abstracts broadly applicable principles from cross-persona trajectories. Subsequently, Persona Delta Distillation (PDD) (Section~\ref{sec:pdd}) captures idiosyncratic rules specifically for users exhibiting divergent behaviors. Finally, a Cross-Level Rule Flow mechanism (Section~\ref{sec:flow}) calibrates the partition by migrating rules between the two tiers as evidence accumulates. The overall architecture of HiPS is illustrated in Fig.~\ref{fig:method}.

\subsection{Universal Strategy Distillation (USD)}
\label{sec:usd}
Existing strategies are either manually crafted or learned from data, yet both are subsequently frozen, thereby failing to incorporate dynamic evidence accumulated as the policy evolves. To break this static paradigm, $\Su$ should be continuously refined using live trajectories. While contrastive analysis of trajectories can identify promising patterns, distilling these insights into reliable rule updates presents two challenges: (1) Conventional free-form feedback treats the guideline as a monolithic text, lacking per-rule granularity. (2) Without granular evidence tracking, a rule proposed in one cycle may be arbitrarily overwritten in the next. USD resolves these issues by utilizing \emph{structured diffs} to decompose updates into rule-level operations, coupled with a rigorous evidence tracking mechanism to maintain a persistent validation history.

\paragraph{Contrastive feedback as structured diff.}
At regular training intervals, USD samples a persona-balanced set of the top-$k$ and bottom-$k$ trajectories to prevent dominant users from skewing the optimization signal. An LLM meta-optimizer is prompted to compare these contrastive sets. Instead of yielding a free-form summary, the optimizer outputs a structured diff, a discrete set of operations applied to the current rule set (see Appendix~\ref{sec:prompts} for prompt). Formally,
\begin{equation}
    \delta_{\text{USD}} = \{\text{V}: [\cdot],\ \text{H}: [\cdot],\ \text{R}: [\cdot]\},
\end{equation}
where V signifies validating an existing rule, H represents hypothesizing a new rule, and R denotes revising or retiring a rule. Crucially, new hypotheses must strictly adhere to the ``[Label]: When [condition], [action]'' format and prescribe explicit management behaviors.

\paragraph{Evidence tracking.}
Since a single distillation cycle may yield spurious hypotheses, USD incorporates strict evidence tracking to filter out noise. Each rule's evidence level, including \texttt{supported, established}, and \texttt{tentative} (see Appendix~\ref{sec:evidence} for details) is iteratively updated based on its validation signals: consistent validation operations promote the rule, whereas revision or contradiction operations demote it. To complement the LLM judgment with a data-driven safeguard, rules are automatically promoted if their predictive gain (Eq.~\ref{eq:pg}) exceeds $\theta_{\text{val}}$, and demoted if it falls below $\theta_{\text{rev}}$.

\subsection{Persona Delta Distillation (PDD)}
\label{sec:pdd}

Under globally shared strategies, rules benefiting a minority receive diluted signals while locally harmful ones persist. Although per-user adaptation resolves this, naively generating $\Dp$ for everyone wastes computational budget and introduces noise for those adequately served by $\Su$. To mitigate this, PDD employs a \emph{divergence criterion} to gate personalization, selectively identifying users who genuinely require adaptive rules.

\paragraph{Divergence-gated personalization.}
To quantify behavioral deviation, we first evaluate the extent to which each universal rule correlates with task success. We define the \emph{predictive gain} of a rule $r$ as the absolute difference in the expected success rate (denoted by $Y \in \{0, 1\}$) between trajectories that comply with $r$ and those that do not:
\begin{equation}
    \PG(r) = \big| P(Y\!=\!1 \mid r^+) - P(Y\!=\!1 \mid r^-) \big|,
\label{eq:pg}
\end{equation}
where $r^+$ and $r^-$ indicate rule compliance and non-compliance, respectively. PG provides a proxy for rule importance: rules with high PG strongly predict task outcomes, while near-zero PG signals irrelevance. Crucially, PG is used only for relative ranking and threshold-based gating, not as an absolute measure of rule quality. Building upon this, we define the divergence of user $p$ as the aggregate discrepancy in rule efficacy between the specific user and the broader population:
\begin{equation}
    \Diverg_p = \sum_{r \in \Su} \left| \PG(r \mid p) - \PG(r) \right|.
\label{eq:divergence}
\end{equation}
Users exhibiting a divergence $\mathcal{D}(p) \geq \theta_{\text{div}}$ trigger the PDD module. Conversely, users below this threshold are deemed sufficiently covered by $\Su$ and bypass the personalization step. This gating mechanism resolves the cold-start dilemma by defaulting low-divergence users to the universal baseline, optimizing resource allocation.

\paragraph{Behavior-conditioned distillation.}
For users identified as divergent, PDD prompts an LLM to formulate management behaviors that strictly differ from $\Su$, using the universal rules as an anchoring context. To prevent $\Dp$ from degenerating into mere factual knowledge, the generated rules are constrained to be \emph{management-oriented} to prescribe actions rather than asserting facts and \emph{behavior-conditioned}. This ensures robust generalization to unseen users at inference time.

\subsection{Cross-Level Rule Flow}
\label{sec:flow}

If $\Su$ and $\Dp$ evolve in strict isolation, the boundary between them becomes brittle. Without a dynamic transfer mechanism, a personalized rule that ultimately proves globally beneficial remains permanently trapped in $\Dp$, while a universal rule contradicted by a subset of users persists erroneously in $\Su$. To resolve this structural rigidity, the Cross-Level Rule Flow dynamically calibrates the partition by migrating rules between tiers based on accumulated cross-persona evidence.

\paragraph{Generalization ($\Dp \rightarrow \Su$).}
When a $\Dp$ rule at ``supported'' level or above appears in $\geq \theta_{\text{flow}}$ fraction of personas that have $\Dp$, it is promoted to $\Su$ and pruned from individual deltas. Intuitively, a management pattern emerging across diverse users signifies a universal principle rather than a localized adaptation. To ensure robust aggregation, we match rules via semantic similarity, consolidating equivalent behaviors despite syntactic variations.

\paragraph{Specialization ($\Su \rightarrow \Dp$).}
When USD revises or demotes a rule within $\Su$, users who previously benefited from the original rule risk losing critical guidance. To mitigate this, PDD hypothesizes persona-specific replacements, effectively restoring the management behavior in a localized form. This mechanism prevents a one-size-fits-all revision from deteriorating the performance of minority users who relied on the deprecated rule.

Together, generalization and specialization establish a cycle. Local discoveries that prove widely beneficial are assimilated into $\Su$, whereas universal rules that are contradicted by specific subpopulations are replaced by tailored adaptations in $\Dp$.

\subsection{Strategy Injection and Co-Evolution}
\label{sec:injection}

In this co-evolutionary framework, the active strategy directs policy rollouts, while the resulting trajectories provide empirical feedback for subsequent strategy refinement. We link these phases using budgeted rule injection and a combined reward that decouples strategy evolution from adherence.

\paragraph{Submodular selection.}
The active strategy $\Sp = \Su \cup \Dp$ is constrained by a maximum token budget $B$. To select the most representative and diverse set of rules, we greedily maximize a submodular coverage objective penalized by redundancy:
\begin{equation}
    \max_{\substack{S \subseteq \Sp \\ \sum_{r \in S} |r| \leq B}} \sum_{r \in S} \PG(r) \!\cdot\! \big(1 - \max_{s \neq r} \sigma(r, s)\big),
\label{eq:submodular}
\end{equation}
where $\sigma(\cdot, \cdot)$ denotes semantic similarity. The token budget is allocated between the universal and personalized levels proportionally to their aggregate predictive gains, subject to a minimum floor for each level as detailed in Appendix~\ref{sec:hyperparams}.

\paragraph{Guideline-aligned reward.}
To align the policy with the discovered strategies, we define a persona-aware adherence reward:
\begin{equation}
    \Rfollow(\tau, p) = \frac{\sum_{r \in \Sp} \PG(r) \cdot h(\tau, r)}{\sum_{r \in \Sp} \PG(r)} \label{eq:rfollow},
\end{equation}
where $h(\tau, r) \in [0,1]$ is a compliance check. The combined reward for Group Relative Policy Optimization (GRPO)~\cite{grpo} is $R(\tau, p) = \Rans(\tau) + \lambda \cdot \Rfollow(\tau, p)$. 
Consequently, the identical trajectory naturally yields varying adherence scores depending on the user's specific $\Dp$.

\paragraph{Anti-circularity and co-evolution.}
A rule could validate itself if compliance inflates the signal used to select which trajectories inform strategy updates. We disrupt this direct path by ranking the distillation buffer exclusively by $\Rans$, while restricting $\Rfollow$ to GRPO advantage computation. An indirect path remains where $\Rfollow$ shapes the policy, which alters rollout behavior and subsequently affects future $\Rans$. However, its indirect path is benign when rule compliance genuinely improves task success, which is precisely the condition under which strategy refinement should occur. A vulnerability arises only when compliance is rewarded without corresponding task improvement. Separating $\Rans$ for distillation from $\Rfollow$ for policy optimization prevents this reinforcement loop. The detailed training procedure, detailed in Appendix~\ref{app:alog}, integrates these components into a unified loop. 

\begin{table*}[!htbp]
\centering
\setlength{\tabcolsep}{3.5pt}
\begin{tabular}{l|cc|cc|cccc|cccc}
\toprule
\multirow{2}{*}{\textbf{Method}} & \multicolumn{2}{c|}{\cellcolor{red!12}\textbf{PersonaMem}} & \multicolumn{2}{c|}{\cellcolor{blue!12}\textbf{PrefEval}} & \multicolumn{4}{c|}{\cellcolor{orange!12}\textbf{PersonaBench}} & \multicolumn{4}{c}{\cellcolor{teal!12}\textbf{PERMA}} \\
& \cellcolor{red!12}32K & \cellcolor{red!12}128K & \cellcolor{blue!12}Expl. & \cellcolor{blue!12}Impl. & \cellcolor{orange!12}0 & \cellcolor{orange!12}0.3 & \cellcolor{orange!12}0.5 & \cellcolor{orange!12}0.7 & \cellcolor{teal!12}C-S & \cellcolor{teal!12}C-M & \cellcolor{teal!12}N-S & \cellcolor{teal!12}N-M \\
\midrule
Long Context & 42.17 & 20.74 & 32.40 & 25.90 & 27.77 & 22.38 & 19.45 & 12.59 & 24.11 & 22.88 & 24.82 & 22.81 \\
RAG          & 52.41 & 38.02 & 47.40 & 32.40 & 32.21 & 29.13 & 25.28 & 23.36 & 51.63 & 36.84 & 51.77 & 35.09 \\
\midrule
Mem0         & 48.53 & 39.67 & 57.60 & 46.40 & 17.43 & 18.95 & 19.31 & 16.49 & 52.76 & 51.67 & 51.49 & 51.15 \\
A-Mem        & 55.42 & 39.88 & 62.30 & 52.80 & 30.09 & 28.52 & 25.81 & 24.10 & 53.05 & 53.73 & 52.91 & 49.87 \\
LightMem     & 52.41 & 36.74 & 64.20 & 54.80 & 19.14 & 17.83 & 19.61 & 17.51 & 59.43 & 55.53 & 59.43 & 55.53 \\
\midrule
MemAgent     & 59.93 & 50.86 & 77.40 & 64.40 & 30.74 & 28.34 & 24.47 & 23.53 & 53.98 & 53.90 & 55.74 & 48.84 \\
Mem-$\alpha$ & 58.43 & 48.18 & 75.70 & 63.10 & 19.31 & 18.01 & 17.35 & 16.64 & 52.19 & 51.67 & 54.46 & 46.53 \\
MemSkill     & 64.45 & 55.37 & 82.60 & 68.40 & 31.19 & \textbf{29.89} & 24.85 & 23.36 & 62.83 & 51.92 & 60.14 & 53.21 \\
\midrule
\rowcolor{rowours} \textbf{HiPS (Ours)} & \textbf{73.49} & \textbf{62.01} & \textbf{89.20} & \textbf{69.40} & \textbf{32.27} & 29.76 & \textbf{25.99} & \textbf{25.09} & \textbf{66.95} & \textbf{56.56} & \textbf{63.83} & \textbf{56.30} \\
\bottomrule
\end{tabular}
\caption{\textbf{Overall comparison across twelve evaluation settings.} PersonaMem (32K/128K) is in-domain; PrefEval (Explicit/Implicit), PersonaBench (4 noise levels), and PERMA (C-S/C-M/N-S/N-M denote Clean/Noise $\times$ Single/Multi-Domain) are out-of-domain. Higher is better. Best results are in \textbf{bold}.}
\label{tab:main}
\end{table*}

\section{Experiments}

\subsection{Experimental Setup}

\paragraph{Benchmarks.}

We evaluate on four personalized memory benchmarks: \textbf{PersonaMem}~\cite{personamem} for preference evolution over varying context scales; \textbf{PrefEval}~\cite{prefeval} for explicit and implicit queries with 50 distraction turns; \textbf{PersonaBench}~\cite{personabench} for personalized retrieval and QA over corpora; and \textbf{PERMA}~\cite{perma} combining two noise levels with two task scopes. We report accuracy for PersonaMem, PrefEval, and PERMA, and the F1 score for PersonaBench. See Appendix~\ref{app:Datasets} for details.

\paragraph{Baselines.}
We compare our framework against a diverse selection of baselines. \textbf{LongContext} feeds raw histories directly into the context window, whereas \textbf{RAG} retrieves the top-$k$ relevant segments from a vector store. We also include three retrieval-based methods that dynamically maintain an external memory bank: \textbf{Mem0}~\cite{mem0}, \textbf{A-Mem}~\cite{amem}, and \textbf{LightMem}~\cite{lightmem}. Furthermore, we compare against three RL-based memory agents that learn memory evolution actions: \textbf{MemAgent}~\cite{memagent2026}, \textbf{MEM-$\alpha$}~\cite{memalpha}, and \textbf{MemSkill}~\cite{memskill}. For a fair comparison, all baselines use the same configurations.

\paragraph{Implementation.}
We employ Qwen2.5-7B-Instruct~\cite{qwen} as the backbone language model for most methods, whereas \textsc{Mem-$\alpha$} utilizes Qwen3-4B and MemSkill uses Qwen3-Next-80B-A3B-Instruct. For retrieval, all-MiniLM-L6-v2~\cite{minilm} extracts the top ten candidates. Our training set consists of 423 instances sampled from 70\% of the PersonaMem 32k subset. To reduce computational overhead, training uses retrieved dialogues as context, whereas inference utilizes the full history, processing context in 4K-token chunks. Baselines are implemented using their public repositories; for a fair comparison, MemAgent, \textsc{Mem-$\alpha$} and MemSkill are initialized from public checkpoints and fine-tuned on our 423 samples. All experiments run on four NVIDIA H800 GPUs, with shared hyperparameters detailed in Appendix~\ref{sec:hyperparams}.

\subsection{Main Results}

\paragraph{Overall Comparison with Baselines.}
As shown in Tab.~\ref{tab:main}, HiPS consistently demonstrates superior performance across all twelve evaluation settings. This performance gap indicates that learning an explicit, dynamically evolving memory-management strategy is significantly more effective than relying on fixed context inclusion or manually designed retrieval heuristics. Specifically, the performance of the Long Context baseline degrades severely under noisy, long-horizon histories, falling to 20.74 on PersonaMem 128K. In contrast, HiPS maintains stable personalized performance by utilizing its distilled strategies to filter out irrelevant information during memory updates. Compared to explicit memory-bank baselines such as Mem0, A-Mem, and LightMem, HiPS delivers substantial and consistent improvements across both in-domain and out-of-domain tasks. Although reinforcement learning-based memory agents, specifically MemAgent, \textsc{Mem-$\alpha$}, and MemSkill, are competitive, they still lag behind the holistic performance of our framework. This consistent advantage highlights the efficacy of our hierarchical co-evolution design, where universal strategy distillation induces a transferable global guideline, and persona delta distillation dynamically adapts to user-specific behavioral patterns.
\paragraph{Generalization Across Settings.}
Across the performance detailed in Tab.~\ref{tab:main}, HiPS exhibits robust generalizability. It consistently outperforms baselines on both in-domain tasks and out-of-domain benchmarks. This includes the explicit and implicit preference queries of PrefEval, the increasingly noisy contexts of PersonaBench, and the diverse domain crossings of PERMA. This strong generalization capability is driven by our hierarchical strategy decomposition, where universal strategy distillation establishes stable, globally shared memory organizations, and persona delta distillation handles targeted adaptations for divergent users. Detailed performance comparisons across different user categories are provided in Appendix~\ref{app:ComparisoninDifferentCategories}.

\begin{table}[t]
\centering
\small
\colorlet{dropA}{red!6}
\colorlet{dropB}{red!15}
\colorlet{dropC}{red!24}
\colorlet{dropD}{red!33}
\colorlet{dropE}{red!42}
\setlength{\tabcolsep}{3.1pt}
\begin{tabular}{@{}l cc cccc@{}}
\toprule
& \multicolumn{2}{c}{\textbf{PersonaMem}} & \multicolumn{4}{c}{\textbf{PERMA}} \\
\cmidrule(lr){2-3} \cmidrule(lr){4-7}
\textbf{Configuration} & 32K & 128K & C-S & C-M & N-S & N-M \\
\midrule
\rowcolor{rowours} HiPS (Full) & \textbf{73.49} & \textbf{62.01} & \textbf{66.95} & \textbf{56.56} & \textbf{63.83} & \textbf{56.30} \\
w/o Flow     & \cellcolor{dropA}71.08 & \cellcolor{dropA}60.49 & \cellcolor{dropE}45.39 & \cellcolor{dropE}40.87 & \cellcolor{dropE}43.12 & \cellcolor{dropE}36.50 \\
w/o Gate     & \cellcolor{dropB}69.28 & \cellcolor{dropA}59.63 & \cellcolor{dropE}51.63 & \cellcolor{dropD}44.99 & \cellcolor{dropD}52.62 & \cellcolor{dropC}46.79 \\
w/o USD      & \cellcolor{dropC}66.27 & \cellcolor{dropC}53.70 & \cellcolor{dropB}61.28 & \cellcolor{dropB}51.93 & \cellcolor{dropB}60.14 & \cellcolor{dropC}49.10 \\
w/o PG       & \cellcolor{dropC}67.47 & \cellcolor{dropC}52.22 & \cellcolor{dropC}60.70 & \cellcolor{dropB}51.41 & \cellcolor{dropB}59.43 & \cellcolor{dropA}53.98 \\
w/o PDD      & \cellcolor{dropB}69.88 & \cellcolor{dropB}56.05 & \cellcolor{dropB}60.99 & \cellcolor{dropA}53.98 & \cellcolor{dropB}60.71 & \cellcolor{dropB}51.41 \\
\bottomrule
\end{tabular}
\caption{Ablation results on PersonaMem and PERMA. Cell color intensity indicates drop from Full (\textcolor{dropA}{\rule{0.8em}{0.8em}}~$<$3 to \textcolor{dropE}{\rule{0.8em}{0.8em}}~$\geq$15\,pp). Rows ordered by average drop.}
\label{tab:ablation}
\end{table}

\subsection{Ablation Study}

Tab.~\ref{tab:ablation} ablation results reveal a striking in-domain versus out-of-domain importance flip between PersonaMem and PERMA. In-domain, removing \textbf{USD} or \textbf{PG} causes the largest declines, reducing PersonaMem 128K accuracy from 62.01 to 53.70 and 52.22, respectively, confirming that universal rule quality is critical when distributions align. Under this setting, \textbf{PDD} yields a moderate drop to 56.05, while \textbf{Gate} and \textbf{Flow} reduce accuracy only to 59.63 and 60.49, as training personas are well-served by $\Su$ alone. Conversely, this ranking reverses out-of-domain. Ablating \textbf{Flow} triggers the most severe degradation, reducing C-S accuracy from 66.95 to 45.39, an average PERMA decrease of 19.4 points, while removing \textbf{Gate} drops C-S accuracy to 51.63, averaging an 11.9-point loss. These drops show that cross-level rule migration and divergence-gated personalization are essential when universal rules do not transfer, especially in multi-domain settings such as C-M and N-M, where cross-domain interests stress-test persona diversity. In contrast, USD and PG matter less out-of-domain, dropping performance by only 5.3 and 4.5 points on average. Finally, ablating \textbf{PDD} causes consistent, moderate drops of 6.0 on PersonaMem 128K and 4.1 on average across PERMA.

\begin{figure}[t]
\centering
\includegraphics[width=\columnwidth]{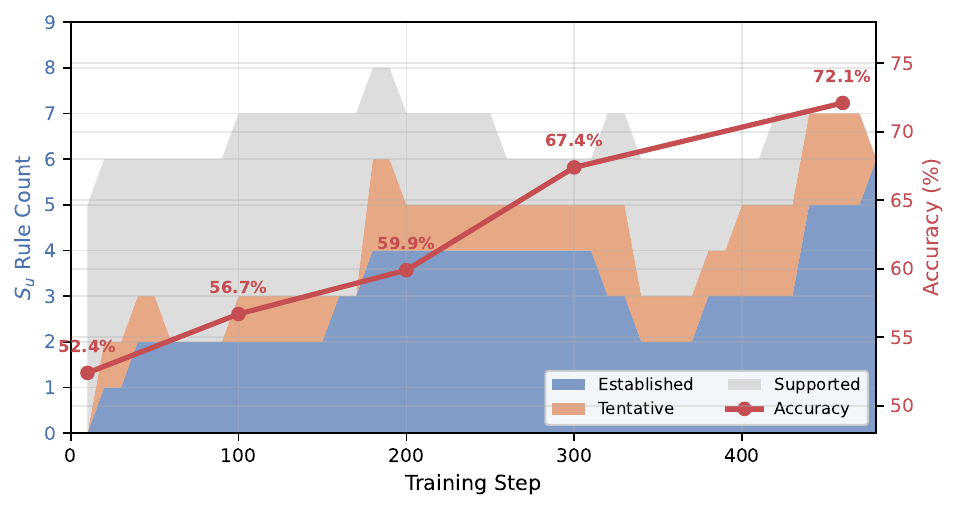}
\caption{Evolution of $S_u$ evidence distribution and accuracy throughout training on PersonaMem.}
\vspace{-0.25cm}
\label{fig:strategy-evolution}
\end{figure}

\subsection{Strategy Quality Analysis}

\paragraph{Evolutionary Dynamics of Universal Strategies.}
 Fig.~\ref{fig:strategy-evolution} traces $\Su$ evidence distribution and accuracy throughout training on PersonaMem 32K validation set. Starting from 5 seed rules (all supported), the system hypothesizes new rules (appearing as tentative) and validates successful ones. The stacked area shows evidence of evolution: rules progressively mature from tentative to supported to established, with accuracy rising from 52.4\% to 72.1\%  by convergence. The rule count oscillates before stabilizing at 6 established rules at step 480+, reflecting the create-validate-prune lifecycle. 

\begin{figure*}[t]
    \centering
    \includegraphics[width=0.99\linewidth]{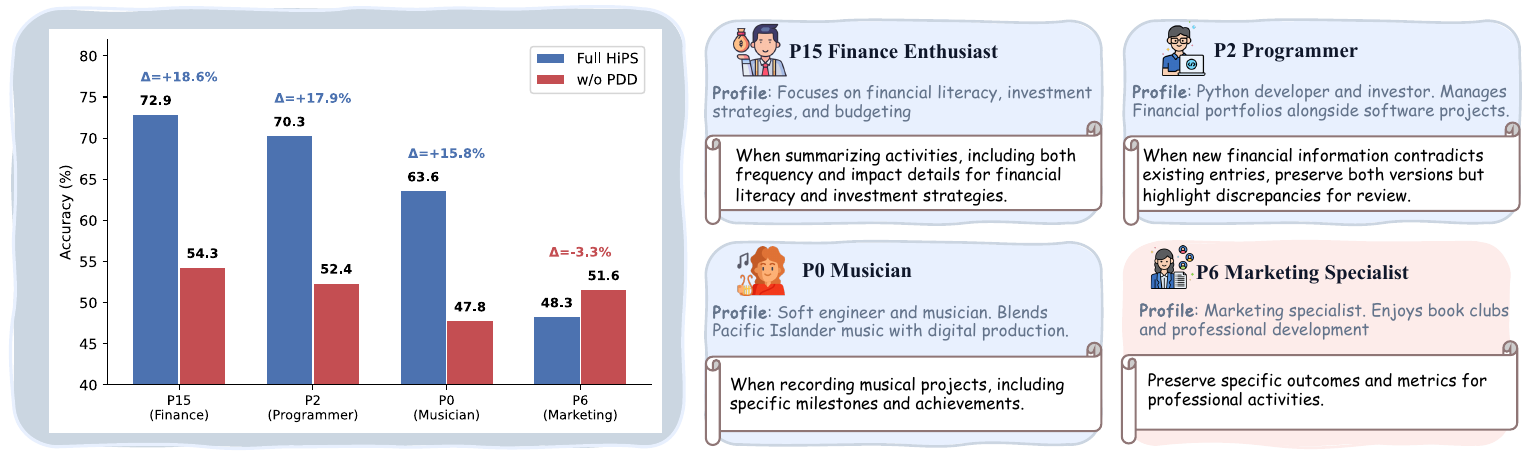}
    \caption{Per-persona impact of persona-specific strategies ($\Delta_p$). We compare Full HiPS (with $\Delta_p$) against w/o PDD (without $\Delta_p$) on PersonaMem 128K. PDD yields an average improvement of +6.0\%, benefiting 20 personas. However, the gains are highly heterogeneous: P15 (finance enthusiast) gains +18.6\%, while P6 (marketing specialist) suffers -3.3\%. We highlight four representative personas to illustrate why.}
    \label{fig:case}
\end{figure*}

\paragraph{Qualitative Analysis of Personalized Deltas.}
Fig.~\ref{fig:case} illustrates how the impact of $\Dp$ varies across user profiles. For P15, the finance enthusiast, a specialized financial logging rule yields an 18.6\% point improvement by preventing granular data compression. Similarly, P2, the programmer, gains 17.9\% points by preserving conflicting portfolio entries instead of overwriting them, while P0, the musician, gains 15.8\% points through custom milestone tracking. Conversely, P6, the marketing specialist, experiences a 3.3\% point drop. Because P6's professional activities are already well-managed by $\Su$, additional rules over-constrain model generation, validating our divergence-gating design to selectively apply personalization.

\begin{table}[t]
  \centering
  \small
  \setlength{\tabcolsep}{1.9pt}
  \begin{tabular}{l|c c c c}
  \toprule
  \textbf{Method} & \makecell{\textbf{Qwen2.5-7B}\\\textbf{Instruct}}
  & \makecell{\textbf{GPT-4o}\\\textbf{-mini}}
  & \makecell{\textbf{Gemini 2.5}\\\textbf{flash}}
  & \textbf{GPT-5} \\
  \midrule
  RAG & 52.41 & 51.58 & 60.83 & 63.74 \\
  A-Mem & 55.42 & 55.91 & 62.74 & 65.93 \\
  \midrule
   \multicolumn{5}{l}{\cellcolor{cvprblue!7}\textit{$\blacktriangledown$ Optimized w/ Qwen2.5-7B-Instruct}} \\
  HiPS ($S_u$ only) & 57.83 & 56.74 & 64.25 & 67.18 \\
  HiPS ($S_u$+$\Delta_p$) & \textbf{61.24} & 59.83 & 67.45 & 70.16 \\
  \midrule
   \multicolumn{5}{l}{\cellcolor{cvprblue!7}\textit{$\blacktriangledown$ Optimized w/ GPT-4o-mini}} \\
  HiPS ($S_u$ only) & 56.47 & 57.35 & 65.18 & 68.42 \\
  HiPS ($S_u$+$\Delta_p$) & 60.07 & \textbf{61.35} & \textbf{68.12} & \textbf{71.48} \\
  \bottomrule
  \end{tabular}
  \caption{Cross-model transferability of strategies (without RL policy). Strategies are distilled with one LLM and injected into others' prompts for inference.}
  \label{tab:cross_llm}
  \end{table}

\subsection{Cross-model Transfer}

Tab.~\ref{tab:cross_llm} presents the transferability of the distilled strategies across various backbone LLMs. Both HiPS variants consistently outperform baselines such as RAG and A-Mem across all models, demonstrating that our strategy captures model-agnostic memory management principles rather than overfitting to a specific LLM. Moreover, the $S_u$+$\Dp$ configuration surpasses the $S_u$-only baseline, confirming that hierarchical personalization provides complementary benefits across backbones. Notably, optimization via \texttt{GPT-4o-mini} generalizes well, achieving strong results on three backbones, including \texttt{GPT-5} at 71.48 and \texttt{Gemini 2.5 flash} at 68.12. These results confirm that HiPS produces portable strategies, enabling deployment across diverse backbone models.

\begin{figure}
    \centering
    \includegraphics[width=\linewidth]{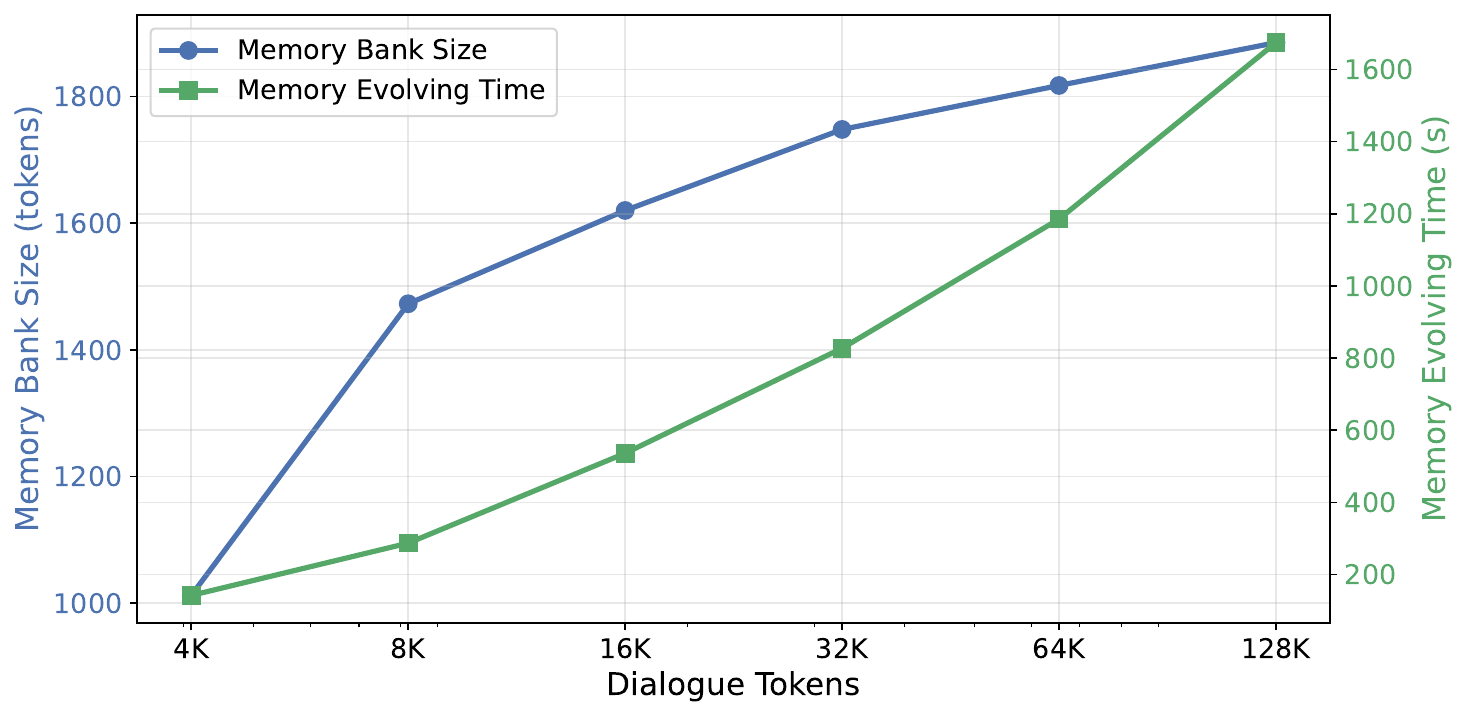}
    \caption{Scaling analysis on PersonaMem (128K). We increase the total dialogue tokens (each evolution round processes 4K tokens) and report the resulting memory bank size (left) and memory evolving time (right).}
    \vspace{-0.25cm}
    \label{fig:scaling}
\end{figure}
\subsection{Scaling Analysis}
 Fig.~\ref{fig:scaling} analyzes how our framework scales with longer dialogue context on PersonaMem (128K). As the dialogue tokens grow from 4K to 128K, the memory bank size increases from 1,000 to around 1,900 tokens, and the curve exhibits a clear trend: it rises steadily in the short-context regime and gradually flattens as the dialogue lengthens, which demonstrates that HiPS effectively consolidates preference-relevant information while filtering out redundant content to keep memory growth bounded. Meanwhile, the memory evolving time scales approximately linearly with dialogue length, ranging from roughly 140 to 1,700 seconds across the full spectrum. This indicates that the computational overhead remains predictable and well-controlled, confirming that HiPS can efficiently handle ultra-long dialogues without incurring prohibitive memory or time costs.

\section{Conclusion}

In this paper, we introduce HiPS, a framework that decouples memory management strategies into a universal baseline and a user-specific adaptive delta, with the partition updated using online evidence. Our empirical results verify that HiPS secures consistent performance gains over retrieval and RL-trained counterparts, with its advantages becoming pronounced as the dialogue context scales. By establishing a continuous co-evolution loop between multi-tiered strategies and active policies, this work paves the way for developing robust and truly personalized long-term memory systems for agents.

\section*{Limitations}

While HiPS demonstrates robust improvements across diverse memory benchmarks, we identify several gentle limitations that offer promising avenues for future research. First, our evaluation is primarily focused on English-language conversational benchmarks. While the core algorithmic principles of strategy distillation and policy co-evolution are model-agnostic, memory management strategies in other typological languages may exhibit different syntactic structures or require localized prompt adjustments. Testing our framework in multilingual and cross-lingual settings remains an open area for future work. Second, our experiments evaluate interaction histories spanning up to 1M tokens, which represents a medium-to-long timeline of conversational sessions. Over extremely long-term lifecycles, such as years of continuous daily interaction, a user's fundamental baseline personality traits may undergo gradual, paradigm-level shifts. Handling such slow-moving, long-term personal baseline drift would require additional meta-distillation mechanisms to update the universal seed rules themselves occasionally.

\section*{Acknowledgments}

This research was partially supported by the National Natural Science Foundation of China (Grants No.62406303).

\bibliography{references}

\appendix

\section{Datasets} 
\label{app:Datasets}

\subsection{PersonaMem Benchmark}
\label{app:personamem-dataset}

PersonaMem \cite{personamem} is a large-scale benchmark designed to evaluate long-term personalization in conversational LLMs. It features interaction histories for \textbf{20 simulated personas}, with each persona defined by rich static attributes, such as demographics, alongside dynamic traits and preferences that evolve across \textbf{15 diverse real-world task domains}, including food recommendation, travel planning, and therapy consultation. For every persona, multi-session conversations are constructed where the user engages with a chatbot via \textbf{7 types of in-situ queries} that probe distinct personalization capabilities, such as recalling user facts, tracking preference evolution, and providing preference-aligned suggestions. Each session consists of 15 to 30 user-assistant turns. These histories are instantiated at three context scales by concatenating 10, 20, or 60 sessions, yielding approximate context lengths of 32k, 128k, and 1M tokens, respectively. At evaluation time, models must select appropriate responses to user queries conditioned on the interaction history, testing their ability to adapt to dynamic user profiles. Tab.~\ref{tab:personamem-stats} summarizes the main statistics of PersonaMem.

\begin{table}[t]
\setlength{\tabcolsep}{4pt}
  \centering
  \begin{tabular}{lccc}
    \toprule
    \textbf{Statistic} & \multicolumn{3}{c}{\textbf{PersonaMem}} \\
    \midrule
    Tokens per history        & \textasciitilde 32k   & \textasciitilde 128k  & \textasciitilde 1M     \\
    \# QA pairs               & 589   & 2727  & 2674   \\
    \# Sessions per history   & 10    & 20    & 60     \\
    Avg.\ \# utterances       & 167.1 & 758.3 & 3607.9 \\
    \bottomrule
  \end{tabular}
  \caption{Statistics of the PersonaMem dataset at different context lengths. Token counts denote the approximate total context length per interaction history; utterance counts are averaged over histories.}
  \label{tab:personamem-stats}
\end{table}

\begin{table}[t]
\centering
\begin{tabular}{l r}
\toprule
\textbf{Statistic} & \textbf{Value} \\
\midrule
Explicit queries & 1{,}000 \\
Implicit queries & 1{,}000 \\
Maximum inserted conversations & 24 \\
Maximum inserted turns & 326 \\
Avg.\ turns / conversation & 13.58 \\
Total tokens & 108{,}102 \\
Avg.\ tokens / conversation & 4{,}504.25 \\
\bottomrule
\end{tabular}
\caption{Dataset statistics for PrefEval multiple-choice classification.}
\label{tab:prefeval_stats}
\end{table}

\begin{table}[t]
\centering
\small
\begin{tabular}{lrrrrr}
\toprule
\textbf{User} & \textbf{Queries} & \textbf{Corpus} & \textbf{Conv.} & \textbf{AI} & \textbf{E-com.} \\
\midrule
1 & 48 & 110 & 84 & 23 & 3 \\
2 & 43 & 90  & 78 & 8  & 4 \\
3 & 42 & 64  & 51 & 12 & 1 \\
4 & 46 & 85  & 71 & 14 & 0 \\
5 & 44 & 84  & 59 & 21 & 4 \\
6 & 40 & 94  & 79 & 14 & 1 \\
\midrule
\textbf{Sum} & 263 & 527 & 422 & 92 & 13 \\
\bottomrule
\end{tabular}
\caption{Statistics of the PersonaBench subset across six users. \textit{Corpus} is the sum of \textit{Conv.}, \textit{AI}, and \textit{E-com.}.}
\label{tab:personabench-stats}
\end{table}

\subsection{PrefEval Benchmark}
PrefEval~\cite{prefeval} is a long-context, multi-session benchmark designed to evaluate whether LLMs can infer, retrieve, and act on user preferences in realistic conversational settings. The benchmark emphasizes four core aspects: preference inference, long-context retrieval, preference following, and personalization proactiveness. The dataset comprises 1{,}000 unique preference-query pairs and spans 20 everyday topics grouped into seven domains: \textit{Entertainment}, comprising shows, music and books, sports, and games; \textit{Travel}, spanning activities, restaurants, hotels, and transportation; \textit{Lifestyle}, including diet, beauty, fitness, and health; \textit{Shopping}, encompassing home, fashion, motors, and technology; \textit{Education}, covering educational resources and learning styles; \textit{Professional Ownership}; and \textit{Professional Work Style}. PrefEval supports two evaluation formats, namely a free-form generation setting and a 4-way multiple-choice classification setting where exactly one option is consistent with the stated preference. To challenge long-range personalization capabilities, the benchmark inserts unrelated multi-session dialogue turns between the preference revelation and the final query. In our experiments, we employ 1{,}000 explicit and 1{,}000 implicit instances under the multiple-choice classification setting, with 50 intervening turns inserted as distractor context. Tab.~\ref{tab:prefeval_stats} reports the summary statistics of this subset.

\subsection{PersonaBench Benchmark}

PersonaBench~\cite{personabench} evaluates personalized retrieval and question answering grounded in user-specific contexts. For each user, the dataset provides a heterogeneous personal corpus consisting of conversations with friends, denoted as \textit{Conv.}, dialogues with AI assistants, denoted as \textit{AI}, and e-commerce purchase histories, denoted as \textit{E-com.}. The evaluation queries are typically short and underspecified, requiring models to resolve implicit intent by grounding responses in evidence distributed across historical interactions and behaviors. This setup tests a model's ability to align with diverse, user-dependent semantics under realistic contextual ambiguity. Tab.~\ref{tab:personabench-stats} summarizes the per-user query counts and corpus statistics for the six-user subset used in our experiments.

\begin{table*}[htbp]
\centering
\small
\setlength{\tabcolsep}{3pt}
\begin{tabular}{lccccccccccc}
\toprule
\textbf{Metric} & \textbf{P1} & \textbf{P2} & \textbf{P3} & \textbf{P4} & \textbf{P5} & \textbf{P6} & \textbf{P7} & \textbf{P8} & \textbf{P9} & \textbf{P10} & \textbf{Total} \\
\midrule
\multicolumn{12}{l}{\textit{\textbf{User Demographics}}} \\
\midrule
Age & 35-44 & 25-34 & 65+ & 55-64 & 35-44 & 18-24 & 25-34 & 35-44 & 25-34 & 25-34 & -- \\
Gender & M & M & F & M & M & M & M & M & F & M & -- \\
Education & Grad. & Univ. & -- & Univ. & -- & Grad. & Univ. & -- & Grad. & Univ. & -- \\
Country & CA & MX & FI & US & AU & UK & CH & IL & RU & BE & -- \\
\midrule
\multicolumn{12}{l}{\textit{\textbf{Interaction Statistics (Base Dataset: Clean / Noisy)}}} \\
\midrule
\# Interests & 17 & 16 & 15 & 15 & 17 & 13 & 13 & 17 & 16 & 12 & 151 \\
\# Queries & 63 & 62 & 60 & 59 & 63 & 51 & 50 & 63 & 60 & 49 & 580 \\
\# Events & 81 & 81 & 85 & 79 & 82 & 79 & 75 & 81 & 81 & 84 & 808 \\
\# Dialogs & 356/364 & 340/358 & 373/379 & 347/352 & 348/370 & 349/357 & 321/329 & 350/364 & 359/362 & 365/375 & 3.5k/3.6k \\
\# Tokens (k) & 34/34 & 32/33 & 33/36 & 31/31 & 33/33 & 32/32 & 31/31 & 32/36 & 34/39 & 33/34 & 324/331 \\
\midrule
\multicolumn{12}{l}{\textit{\textbf{Style-aligned Long-Context Dataset}}} \\
\midrule
\# Events & 156 & 141 & 156 & 139 & 137 & 107 & 176 & 94 & 155 & 127 & 1,388 \\
\# Dialogs & 1009 & 848 & 1084 & 876 & 888 & 802 & 1015 & 798 & 969 & 869 & 9,158 \\
\# Tokens (k) & 139.0 & 61.4 & 174.0 & 88.9 & 197.2 & 69.0 & 151.1 & 62.7 & 143.7 & 78.4 & 1,165.4 \\
\bottomrule
\end{tabular}
\caption{Comprehensive statistics of the dataset across multiple dimensions. The table summarizes user demographics for profiles, followed by statistics for the base (comparing Clean and Noise) and the style-aligned long-context dataset. Note: For User Demographics Profile (P), Gender: M (Male), F (Female); Education: Grad. (Graduate Degree), Univ. (University); Country codes follow ISO 3166-1 alpha-2. In Interaction Statistics, \# denotes the count, and (k) represents thousands of tokens.}
\label{tab:perma_statistics}
\end{table*}

\subsection{PERMA Benchmark}

PERMA~\cite{perma} is an event-driven, long-context benchmark for evaluating whether personalized memory agents can maintain, update, and synthesize dynamic persona states in realistic conversational environments, with an emphasis on three aspects: task completion, preference consistency, and informational confidence. The dataset comprises 10 representative user profiles containing 2{,}166 fine-grained preference details across 10 countries, and spans 20 distinct domains, including \textit{Travel}, \textit{Finance}, \textit{Shopping}, \textit{Entertainment}, \textit{Messaging}, and \textit{Calendar}. PERMA supports two evaluation formats: an 8-option multiple-choice question (MCQ) probing setting in which options are systematically ablated, and a multi-turn interactive setting driven by an LLM-based user simulator that terminates upon successful task completion. To stress realistic, long-horizon personalization, the benchmark structures interactions temporally through event-driven timelines and injects controlled within-session noise to simulate real-world user erraticism. In our experiments, we evaluate memory agents across four key scenarios--\textit{Clean-Single}, \textit{Clean-Multi}, \textit{Noise-Single}, and \textit{Noise-Multi}--which span single- and multi-domain tasks under both standard dialogue histories and those perturbed with text-variability noise; comprehensive statistics of the dataset are reported in Tab.~\ref{tab:perma_statistics}.

\section{Comparison in Different Categories}
\label{app:ComparisoninDifferentCategories}
\paragraph{Comparison in Different Categories of PersonaMem.}

Tab.~\ref{tab:personamem_comparison} reports category-wise results on PersonaMem under 32K and 128K interaction histories. Across both scales, \ourmodel achieves the best overall performance (73.49 at 32K; 62.01 at 128K), and the gains are concentrated on memory-dependent personalization abilities. On 32K histories, \ourmodel leads in Suggest ideas (51.85), Prefs evolve (84.44), Update reasons (89.66), Aligned recs (73.33), and New Scenarios (53.33), which jointly drives a clear margin over the strongest baselines (e.g., 73.49 vs.\ 64.45 for MemSkill). When scaling to 128K, \textit{Long Context} degrades sharply overall (20.74), while memory-based methods remain substantially stronger; within them, \ourmodel stays best-performing and ranks first on Recall facts (68.42), Suggest ideas (44.02), Latest prefs (71.82), Prefs evolve (71.55), Update reasons (72.49), Aligned recs (54.73), and New Scenarios (44.13). In contrast, Recall facts is not a comparative strength for \ourmodel under the 32K scale (71.43 vs.\ 72.87 for MemAgent), indicating that while simpler fact-retrieval mechanisms can be effective in shorter horizons, \ourmodel's architectural advantage lies in tracking and applying evolving, structured preferences as the context scales up. Overall, the category-wise gains suggest that explicitly \emph{structuring} memory operations and then \emph{selecting} what to keep is most effective for preference-heavy queries, and the advantage becomes more pronounced as the interaction history grows longer.

\paragraph{Comparison in Different Categories of PrefEval.}

Tab.~\ref{tab:prefeval_category} breaks down PrefEval performance by domain under \textbf{Explicit} and \textbf{Implicit} preference settings. Under Explicit Memory, \ourmodel attains the best overall accuracy (89.20) and shows consistently strong gains on preference-heavy domains, ranking first across all evaluated topics, including Travel (90.20), Entertain (94.20), Lifestyle (91.70), Shop (84.10), Education (86.30), Professional (85.20), and Pet (78.10). This comprehensive lead demonstrates that \ourmodel can robustly extract and apply directly stated constraints. Under Implicit Memory, the task becomes more challenging for all methods, yet \ourmodel again leads overall (69.40) and improves most clearly on domains that require inferring latent preferences from context, such as Travel (71.90), Lifestyle (71.60), Shop (64.50), and Professional (63.40). A notable exception is Education, where \ourmodel (65.10) slightly trails MemSkill (65.80), suggesting that certain implicit reasoning contexts may occasionally benefit from alternative skill-driven memory-update pipelines. Compared with memory-bank baselines (Mem0/A-Mem/LightMem), the advantage of \ourmodel is broad across domains in both settings, indicating that it better resists long-range distractors and preserves preference-relevant signals. Overall, these domain-wise results reflect PrefEval’s construction: the inserted unrelated turns make long-range preference retrieval and faithful preference following the main bottlenecks, and \ourmodel improves most on domains where precise preference identification and consistent application are essential.
\begin{table*}[ht]
\centering
\small
\setlength{\tabcolsep}{3pt}
\begin{tabular}{p{2.2cm}|M{1.4cm}M{1.4cm}M{1.4cm}M{1.4cm}M{1.4cm}M{1.4cm}M{1.4cm}|M{1.4cm}}
\toprule
\textbf{Method}  &
\textbf{Recall facts} &
\textbf{Suggest ideas} &
\textbf{Latest prefs} &
\textbf{Prefs evolve} &
\textbf{Update reasons} &
\textbf{Aligned recs} &
\textbf{New Scenarios} &
\textbf{Overall} \\
\midrule
\multicolumn{9}{c}{\textit{\textbf{$\blacktriangledown$ 32K memory corpus data}}} \\
\midrule
Long Context & 41.38 & 22.22 & - & 42.22 & 72.41 & 33.33 & 26.67 & 42.17 \\
RAG          & 37.93 & 22.22 & - & 64.44 & 86.21 & 60.00 & 33.33 & 52.41 \\
Mem0         & 47.93 & 19.41 & - & 46.61 & 79.58 & 57.04 & 42.75 & 48.53 \\
A-Mem        & 45.71 & 18.52 & - & 73.33 & 79.31 & 66.67 & 33.33 & 55.42 \\
LightMem     & 47.06 & 12.90 & - & 66.19 & 75.76 & 38.18 & 24.56 & 52.41 \\
MemAgent     & \textbf{72.87} & 23.66 & - & 69.78 & 83.84 & 50.91 & 38.60 & 59.93 \\
Mem-$\alpha$ & 45.71 & 29.63 & - & 71.11 & 86.21 & 66.67 & 40.00 & 58.43 \\
MemSkill     & 57.14 & 40.74 & - & 73.33 & \textbf{89.66} & 66.67 & 46.67 & 64.45 \\
HiPS         & 71.43 & \textbf{51.85} & - & \textbf{84.44} & \textbf{89.66} & \textbf{73.33} & \textbf{53.33} & \textbf{73.49} \\
\midrule
\multicolumn{9}{c}{\textit{\textbf{$\blacktriangledown$ 128K memory corpus data}}} \\
\midrule
Long Context & 9.26 & 22.15 & 15.08 & 28.87 & 36.49 & 22.02 & 16.67 & 20.74 \\
RAG          & 53.70 & 20.25 & 33.73 & 52.58 & 59.46 & 39.45 & 36.36 & 38.02 \\
Mem0         & 56.25 & 20.49 & 37.77 & 55.09 & 57.20 & 41.64 & 29.13 & 39.67 \\
A-Mem        & 64.81 & 18.99 & 35.71 & 54.64 & 62.16 & 40.37 & 37.88 & 39.88 \\
LightMem     & 30.99 & 14.67 & 35.33 & 61.29 & 64.31 & 35.82 & 28.17 & 36.74 \\
MemAgent     & 56.14 & 27.80 & 61.66 & 64.22 & 66.17 & 40.69 & 34.74 & 50.86 \\
Mem-$\alpha$ & 59.26 & 30.38 & 50.40 & 60.82 & 64.86 & 45.87 & 39.39 & 48.18 \\
MemSkill     & 64.81 & 37.34 & 61.51 & 67.01 & 68.92 & 50.46 & 42.42 & 55.37 \\
HiPS         & \textbf{68.42} & \textbf{44.02} & \textbf{71.82} & \textbf{71.55} & \textbf{72.49} & \textbf{54.73} & \textbf{44.13} & \textbf{62.01} \\
\bottomrule
\end{tabular}
\caption{Category-wise accuracy (\%) on PersonaMem under 32K and 128K interaction histories. “--” indicates the category is not available in the dataset.}
\label{tab:personamem_comparison}
\end{table*}

 \begin{table*}[!htbp]
  \centering
  \small
  \setlength{\tabcolsep}{3pt}
  \begin{tabular}{p{2.2cm}|M{1.4cm}M{1.4cm}M{1.4cm}M{1.4cm}M{1.4cm}M{1.4cm}M{1.4cm}|M{1.4cm}}
  \toprule
  \textbf{Method}  &
  \textbf{Travel} &
  \textbf{Entertain} &
  \textbf{Lifestyle} &
  \textbf{Shop} &
  \textbf{Education} &
  \textbf{Professional} &
  \textbf{Pet} &
  \textbf{Overall} \\
  \midrule
  \multicolumn{9}{c}{\textit{\textbf{$\blacktriangledown$ Explicit Memory}}} \\
  \midrule
  Long Context & 33.90 & 42.30 & 37.80 & 24.10 & 24.70 & 23.70 & 8.00 & 32.40 \\
  RAG & 48.50 & 55.90 & 52.40 & 40.30 & 40.60 & 38.00 & 25.90 & 47.40 \\
  Mem0 & 60.10 & 64.00 & 62.30 & 49.70 & 53.20 & 50.20 & 38.70 & 57.60 \\
  A-Mem & 63.90 & 69.20 & 65.60 & 56.70 & 56.80 & 55.50 & 43.30 & 62.30 \\
  LightMem & 66.10 & 70.90 & 67.20 & 59.30 & 57.50 & 56.30 & 47.00 & 64.20 \\
  MemAgent & 78.60 & 81.10 & 82.40 & 71.60 & 73.20 & 73.10 & 65.20 & 77.40 \\
  Mem-$\alpha$ & 78.10 & 80.20 & 79.40 & 69.80 & 71.60 & 70.70 & 60.60 & 75.70 \\
  MemSkill & 84.10 & 86.50 & 85.00 & 79.10 & 79.30 & 78.50 & 68.80 & 82.60 \\
\textbf{HiPS} & \textbf{90.20} & \textbf{94.20} & \textbf{91.70} & \textbf{84.10} & \textbf{86.30} & \textbf{85.20} & \textbf{78.10} & \textbf{89.20} \\
  \midrule
  \multicolumn{9}{c}{\textit{\textbf{$\blacktriangledown$ Implicit Memory}}} \\
  \midrule
  Long Context & 28.00 & 32.50 & 31.10 & 19.30 & 20.30 & 15.60 & 8.00 & 25.90 \\
  RAG & 33.50 & 42.40 & 36.20 & 24.20 & 27.60 & 24.10 & 9.40 & 32.40 \\
  Mem0 & 48.30 & 53.20 & 51.50 & 39.20 & 40.30 & 39.50 & 26.40 & 46.40 \\
  A-Mem & 54.80 & 58.70 & 56.60 & 46.60 & 49.20 & 45.00 & 34.80 & 52.80 \\
  LightMem & 57.50 & 60.40 & 59.60 & 48.60 & 47.50 & 47.50 & 37.30 & 54.80 \\
  MemAgent & 67.10 & 69.00 & 67.50 & 58.40 & 60.70 & 59.00 & 50.30 & 64.40 \\
  Mem-$\alpha$ & 64.60 & 71.10 & 65.30 & 56.60 & 59.60 & 55.40 & 45.60 & 63.10 \\
  MemSkill & 68.40 & 73.00 & 71.50 & 63.60 & \textbf{65.80} & 62.70 & 54.30 & 68.40 \\
\textbf{HiPS} & \textbf{71.90} & \textbf{74.20} & \textbf{71.60} & \textbf{64.50} & 65.10 & \textbf{63.40} & \textbf{55.30} & \textbf{69.40} \\
  \bottomrule
  \end{tabular}
  \caption{Domain-wise accuracy (\%) on PrefEval multiple-choice classification under Explicit vs.\ Implicit preference.}
  \label{tab:prefeval_category}
  \end{table*}

\begin{table}[t]
\centering
\small
\setlength{\tabcolsep}{8pt} 
\renewcommand{\arraystretch}{1.1}
\begin{tabular}{l c r}
\toprule
\textbf{Hyperparameter} & \textbf{Symbol} & \textbf{Value} \\
\midrule
\multicolumn{3}{l}{\textbf{Universal Strategy Distillation (USD)}} \\
USD distillation frequency & $K_1$ & 20 \\
Distillation trajectories per cycle & $k$ & 5 \\
Trajectory buffer size & — & 50 \\
Promotion threshold & $\theta_{\text{val}}$ & 0.1 \\
Demotion threshold & $\theta_{\text{rev}}$ & 0.02 \\
Stale pruning cycles & $T_{\text{stale}}$ & 3 \\
\midrule
\multicolumn{3}{l}{\textbf{Persona Delta Distillation (PDD)}} \\
PDD distillation frequency & $K_2$ & 10 \\
Gating threshold & $\theta_{\text{div}}$ & 0.3 \\
\midrule
\multicolumn{3}{l}{\textbf{Rule Budget \& Flow}} \\
Total token budget & $B$ & 200 \\
Minimum budget floor & — & $0.2B$ \\
Migration threshold & $\theta_{\text{flow}}$ & 0.6 \\
\midrule
\multicolumn{3}{l}{\textbf{Policy Optimization}} \\
Reward weight (adherence) & $\lambda$ & 0.3 \\
\bottomrule
\end{tabular}
\caption{Hyperparameter settings used for HiPS strategy co-evolution.}
\label{tab:hyperparams}
\end{table}

\section{Hyperparameters}
\label{sec:hyperparams}

\begin{figure*}
    \centering
    \includegraphics[width=\linewidth]{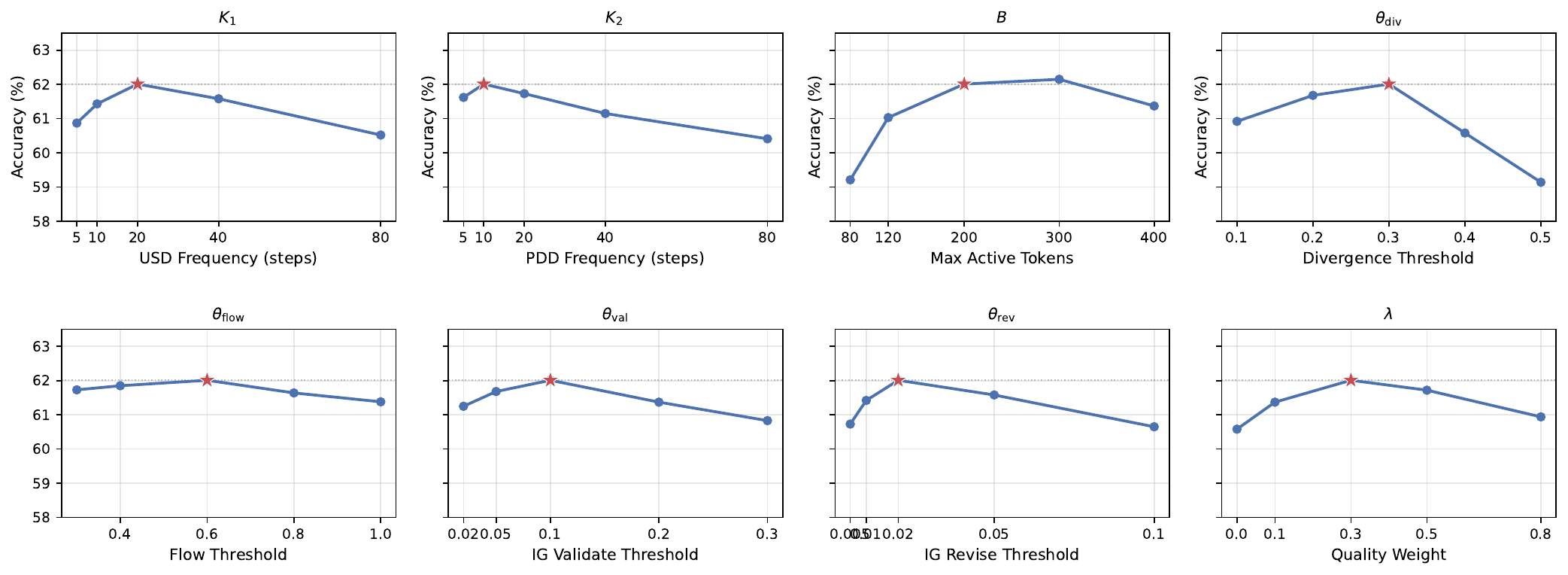}
    \caption{Parameter sensitivity analysis of HiPS on the PersonaMem 128K benchmark. The red stars indicate the selected default values. From top-left to bottom-right, the subplots demonstrate the impact on prediction accuracy of the USD frequency $K_1$, PDD frequency $K_2$, strategy token budget $B$, divergence threshold $\theta_{\text{div}}$, flow threshold $\theta_{\text{flow}}$, validation threshold $\theta_{\text{val}}$, revision threshold $\theta_{\text{rev}}$, and quality weight $\lambda$.}
    \label{fig:hyper}
\end{figure*}
Tab.~\ref{tab:hyperparams} lists all HiPS hyperparameters and their default values.

\paragraph{Hyperparameter Sensitivity Analysis}

To evaluate how key hyperparameters in HiPS influence personalization accuracy, we conduct a parameter sensitivity analysis on the PersonaMem 128K benchmark. As shown in the sensitivity curves, the trends demonstrate clear trade-offs across all key configurations: The distillation frequencies, $K_1$ and $K_2$, govern the update speed of universal and personalized rules. For the USD frequency $K_1$, the optimal performance is achieved at 20 steps. Too frequent distillation, such as every 5 steps, introduces short-term interaction noise, while too infrequent updates, such as every 80 steps, fail to adapt in a timely manner. For the PDD frequency $K_2$, the peak occurs at 10 steps, indicating that personalized strategies require more frequent updates than universal rules to capture rapid user-specific behavioral shifts. Regarding the strategy token budget $B$, performance improves sharply as the budget increases from 80 tokens, peaking around 200 to 300 tokens. Although 300 tokens yields a marginally higher accuracy, we select 200 as the default value to balance strategy expressiveness with prompt token efficiency. The divergence threshold $\theta_{\text{div}}$ controls personalization gating. Low values like 0.1 trigger excessive, unnecessary personalization that introduces content-level noise, whereas high values like 0.5 over-filter users, preventing divergent individuals from receiving tailored rules. The optimal threshold is 0.3. The flow threshold $\theta_{\text{flow}}$ regulates rule migration and exhibits robust performance across a wide range of values, reaching its peak at 0.6. The validation and revision thresholds, $\theta_{\text{val}}$ and $\theta_{\text{rev}}$, peak at 0.1 and 0.02, respectively. These values indicate that a moderately conservative update policy is optimal to prevent the premature promotion of spurious rules while ensuring the timely pruning of contradicted guidelines. Finally, the quality weight $\lambda$ controls the balance between task reward and strategy adherence, peaking at 0.3. Removing the adherence signal entirely, where $\lambda$ is 0.0, reduces performance, while over-weighting adherence at 0.8 degrades task success by forcing compliance over objective outcomes.

\section{Seed Rules for $S_u$}
\label{sec:seeds}

The five seed rules used to initialize $\Su$. All seeds start at the Supported evidence level and must earn Established status through training; they can be revised or pruned if contradicted by evidence.

\begin{enumerate}
    \item \textit{Structure: When processing chunks, organize memory into labeled sections such as Identity, Preferences, and Activities.}
    \item \textit{Conflict resolution: When new information contradicts existing memory, mark the old entry as deprecated with a reason and keep both.}
    \item \textit{Relevance filtering: When encountering one-off details like transient locations or momentary moods, omit them unless explicitly stated as long-term.}
    \item \textit{Compression: When duplicate or redundant entries appear across sections, merge them into a single consolidated entry.}
    \item \textit{Evidence-bound: When updating memory, only store information directly supported by the current conversation chunk.}
\end{enumerate}

\section{Evidence Level Lifecycle}
\label{sec:evidence}

Each rule carries an evidence level that reflects its validation history across distillation cycles:
\begin{itemize}
    \item \textbf{Tentative:} Newly hypothesized, awaiting confirmation. Not injected into the prompt.
    \item \textbf{Supported:} Validated by at least one distillation cycle. Eligible for prompt injection.
    \item \textbf{Established:} Consistently validated across multiple cycles. Highest priority for injection.
\end{itemize}
Rules are promoted when trajectory data shows they predict task success (predictive gain exceeds $\theta_{\text{val}}$), and demoted when contradicted by evidence (predictive gain falls below $\theta_{\text{rev}}$). Stale Tentative rules that receive no validation within $T_{\text{stale}}$ consecutive cycles are pruned.

\section{USD and PDD Prompts}
\label{sec:prompts}
\begin{figure*}[!t]
\centering
\begin{tcolorbox}[
    width=\textwidth,
    colback=gray!5, 
    colframe=gray!50, 
    arc=1mm, 
    boxrule=0.5pt, 
    fonttitle=\bfseries, 
    halign title=center,
    title=Universal Strategy Distillation Template
]
\footnotesize
\begin{verbatim}
You are analyzing a memory management agent that maintains a structured user memory profile across 
multiple conversation sessions.

At each episode, the agent reads conversation chunks one by one, updates a running user memory, and 
then answers a question about the user based solely on this memory.

## Current universal management rules:
{current_rules}

## New evidence from recent episodes across DIFFERENT users:

### High-reward traces, where the memory led to correct answers:
{top_traces}

### Low-reward traces, where the memory led to wrong answers:
{bottom_traces}

## Instructions

Compare the traces. IGNORE user-specific content, such as names, hobbies, or topics. Focus ONLY on how 
the memory was ORGANIZED and MANAGED.

Good rules describe management ACTIONS, for example:
- "Use labeled sections for different information types"
- "Mark changed preferences as deprecated instead of deleting"

Bad rules describe topics, for example:
- "Prioritize cooking preferences", which represents content 
  rather than management

Output ONLY a JSON object with no other text:
- "validate": list of rule indices as integers supported by evidence
- "hypothesize": list of 1 to 2 NEW management rules, requiring fewer than 20 words per rule
- "revise": list of rule indices as integers that conflict with evidence

Example: 
{
 "validate": [0, 2], 
 "hypothesize": ["Use deprecated markers when preferences change"], 
 "revise": [1]
}

If no changes needed: 
{"validate": [], "hypothesize": [], "revise": []}
\end{verbatim}
\end{tcolorbox}
\caption{The complete prompt template used for Universal Strategy Distillation.}
\label{fig:usd-prompt-full}
\end{figure*}

\begin{figure*}[!t]
\centering
\begin{tcolorbox}[
    width=\textwidth,
    colback=gray!5, 
    colframe=gray!50, 
    arc=1mm, 
    boxrule=0.5pt, 
    fonttitle=\bfseries, 
    halign title=center,
    title=Persona Delta Distillation Template
]
\footnotesize
\begin{verbatim}
You are analyzing a memory management agent serving ONE specific user.

The agent already follows these universal management rules:
{universal_rules}

Current personalized rules for this user, where indices refer to this list:
{current_delta}

## New evidence from this user's recent episodes:

### High-reward traces, where the memory worked well:
{top_traces}

### Low-reward traces, where the memory was less useful:
{bottom_traces}

## Instructions

Identify management behaviors specific to THIS user that DIFFER from the universal rules. 
Do NOT repeat any universal rule.

Each rule must describe a MANAGEMENT ACTION, not a topic:
WRONG: "Prioritize financial interests"
RIGHT: "Preserve full preference change history instead of keeping only the latest"

Output ONLY a JSON object with no other text:
- "validate": list of rule indices as integers 
  from personalized rules that are confirmed
- "hypothesize": list of 1 to 2 NEW user-specific 
  management rules, requiring fewer than 20 words per rule
- "revise": list of rule indices as integers 
  that conflict with evidence

Example: 
{"validate": [0], 
 "hypothesize": ["Keep full preference change history with reasons"], 
 "revise": []}

If no changes needed: 
{"validate": [], "hypothesize": [], "revise": []}
\end{verbatim}
\end{tcolorbox}
\caption{The complete prompt template used for Persona Delta Distillation.}
\label{fig:pdd-prompt-full}
\end{figure*}
\paragraph{USD prompt.}
The USD prompt instructs the LLM to compare high-reward and low-reward trajectories across users and identify management patterns that generalize. To focus the LLM on management patterns rather than content, the prompt explicitly instructs: ``IGNORE user-specific content (names, hobbies, topics). Focus only on how the memory was ORGANIZED and MANAGED.'' The prompt is enriched with behavioral feature statistics (verbosity, structural organization, change tracking) and their contrastive attribution across high- and low-reward groups. Key output constraints: (1) structured diffs only (validate/hypothesize/revise), (2) rules follow ``[Label]: When [condition], [action]'' format, (3) rules describe management actions, not topic-level content preferences. When a hypothesized rule overlaps an existing entry (word overlap $> 0.7$), it is treated as a validation rather than a new hypothesis, preventing rule explosion.

\paragraph{PDD prompt.}
The PDD prompt presents $\Su$ as anchoring context and instructs the LLM to identify management behaviors for a specific user that differ from the universal rules. Key constraints beyond USD: (1) rules must describe management actions, not content topics, (2) rule conditions must reference observable behavioral patterns, not user identities or domain names. Concretely, a rule like ``preserve full context for financial discussions rather than summarizing'' is valid because the action is about \emph{how} to handle information; ``the user is interested in finance'' is not, because it states \emph{what} the user cares about without prescribing a management decision. During training, rules are persona-specific; behavior-conditioning enables generalization to unseen users at inference time.

\label{app:alog}
\begin{algorithm}
\caption{HiPS: Hierarchical Strategy-Policy Co-Evolution}
\label{alg:hips}
\begin{algorithmic}[1]
\REQUIRE Policy $\pi_\theta$, seed rules $\Su$, empty $\{\Dp\}$, buffer $\mathcal{B}$
\FOR{each training step $t$}
    \STATE Select $\Sp = \Su \cup \Dp$ via submodular optimization (Eq.~\ref{eq:submodular})
    \STATE Inject $\Sp$ into rollout prompts
    \STATE Collect trajectories $\{\tau\}$; compute $\Rans$, $\Rfollow$
    \STATE Update buffer $\mathcal{B}$ with $(m_K, \Rans, t)$ \hfill \textit{// task reward only}
    \STATE GRPO update on $\pi_\theta$ using $\Rans + \lambda \cdot \Rfollow$
    \IF{$t \bmod K_2 = 0$}
        \STATE Recompute PG for all rules; auto-promote / demote by thresholds
        \STATE PDD: evolve $\Dp$ for users with $\Diverg(p) \geq \theta_{\text{div}}$ (Eq.~\ref{eq:divergence})
    \ENDIF
    \IF{$t \bmod K_1 = 0$}
        \STATE USD: evolve $\Su$ from cross-persona trajectories
        \STATE Generalize frequent $\Dp$ rules $\rightarrow \Su$; consolidate duplicates
    \ENDIF
\ENDFOR
\end{algorithmic}
\end{algorithm}

\section{Compliance Estimation}
\label{sec:compliance}

Both the predictive gain (PG, Eq.~\ref{eq:pg}) and the adherence reward ($R_{\text{follow}}$, Eq.~\ref{eq:rfollow}) require estimating whether a trajectory complies with a given rule. We use a lightweight keyword-based heuristic: for each rule, we define a set of trigger keywords derived from the rule's condition and action clauses; a trajectory segment is marked as compliant ($r^+$) if it contains at least one keyword from the action set and the corresponding condition keywords appear in the preceding context, and as non-compliant ($r^-$) otherwise. This heuristic is intentionally simple to avoid adding LLM calls during training. While imperfect, PG is used for relative ranking (submodular selection, budget allocation) and threshold-based gating (divergence, auto-promotion), both of which are robust to systematic bias in the estimate: a uniform upward shift in all PG values preserves their relative ordering and does not change which users exceed the divergence threshold. The adherence reward $R_{\text{follow}}$ serves as a dense auxiliary signal alongside the sparse task reward $R_{\text{ans}}$; its weight $\lambda = 0.3$ is small, limiting the influence of any single compliance misjudgment.

\section{Algorithm}

Algorithm~\ref{alg:hips} outlines the complete co-evolutionary training procedure for our framework. At each step, the system dynamically constructs an active strategy by merging the universal baseline $\Su$ and the personalized delta $\Dp$ through budgeted submodular selection. This active strategy guides the policy rollouts to generate interaction trajectories. After computing both the task correctness and strategy adherence rewards, the system strictly separates their downstream application. It updates the trajectory distillation buffer using exclusively the task reward to prevent self-validation loops, while utilizing the combined reward to execute the GRPO policy update. To ensure continuous strategy refinement, the framework periodically triggers its dual distillation modules. The system executes Persona Delta Distillation every $K_2$ steps to generate adaptive rules for highly divergent users, and it invokes Universal Strategy Distillation every $K_1$ steps to abstract shared principles and elevate widely successful personalized rules into the global standard.


\end{document}